\pdfoutput=1

\documentclass{cohere}

\usepackage{latexsym}

\usepackage[T1]{fontenc}

\usepackage[utf8]{inputenc}

\usepackage{microtype}

\usepackage{inconsolata}

\usepackage{graphicx}

\usepackage{booktabs} 

\usepackage[ruled,vlined]{algorithm2e}

\usepackage{amsmath}
\usepackage{amsthm}
\usepackage{dsfont}
\usepackage{amssymb}
\usepackage{amscd}
\usepackage{mathtools}
\mathtoolsset{showonlyrefs=true}
\usepackage{enumerate}
\usepackage{ bbold }

\usepackage{xcolor}

\usepackage{tikz}
\usepackage{tcolorbox}
\usepackage{lipsum}

\newlength{\RoundedBoxWidth}
\newsavebox{\GrayRoundedBox}
\newenvironment{GrayBox}[1][\dimexpr\columnwidth-4.5ex]%
   {\setlength{\RoundedBoxWidth}{\dimexpr#1}
    \begin{lrbox}{\GrayRoundedBox}
       \begin{minipage}{\RoundedBoxWidth}}%
   {   \end{minipage}
    \end{lrbox}
    \begin{center}
    \begin{tikzpicture}%
       \draw node[draw=black,fill=black!10,rounded corners,%
             inner sep=2ex,text width=\RoundedBoxWidth]%
             {\usebox{\GrayRoundedBox}};
    \end{tikzpicture}
    \end{center}}

\newcommand{\E}{\mathbb{E}}

\newcommand{\piref}{\pi_\text{ref}}
\newcommand{\x}{\mathbf{x}}
\newcommand{\y}{\mathbf{y}}
\newcommand{\ly}{\mathbf{|y|}}
\newcommand{\lyp}{\mathbf{|y^+|}}
\newcommand{\lym}{\mathbf{|y^-|}}

\title{Averaging log-likelihoods in direct alignment}

\author{
    name = {Nathan Grinsztajn$^\dagger$, Yannis Flet-Berliac$^\dagger$, Mohammad Gheshlaghi Azar, Florian Strub,
    \\
      Bill Wu, Eugene Choi, Chris Cremer, Arash Ahmadian, Yash Chandak, 
    \\
    Olivier Pietquin, Matthieu Geist$^\star$},
    affiliation = {Cohere}
}

\abstract{
To better align Large Language Models (LLMs) with human judgment, Reinforcement Learning from Human Feedback (RLHF) learns a reward model and then optimizes it using regularized RL. Recently, direct alignment methods were introduced to learn such a fine-tuned model directly from a preference dataset without computing a proxy reward function. These methods are built upon contrastive losses involving the log-likelihood of (dis)preferred completions according to the trained model.
However, completions have various lengths, and the log-likelihood is not length-invariant. On the other side, the cross-entropy loss used in supervised training is length-invariant, as batches are typically averaged token-wise.
To reconcile these approaches, we introduce a principled approach for making direct alignment length-invariant. Formally, we introduce a new averaging operator, to be composed with the optimality operator giving the best policy for the underlying RL problem. It translates into averaging the log-likelihood within the loss. 
We empirically study the effect of such averaging, observing a trade-off between the length of generations and their scores.
\looseness=-1
}

\begin{document}

\def\thefootnote{$\dagger$}\footnotetext{Equal contribution.}
\def\thefootnote{$\star$}\footnotetext{Corresponding author: \texttt{matthieu@cohere.com}.}
\def\thefootnote{\arabic{footnote}}

\section{Introduction}

The classic recipe for training a Large Language Model (LLM) is to pre-train the model on a huge dataset, to supervise finetune it from a well-curated dataset of prompt-completion pairs, and to preference finetune it to better align with human judgment. The two first steps rely on minimizing a cross-entropy loss, and are thus invariant to the length of sequences used for training, batches being typically averaged token-wise.

The historical approach uses Reinforcement Learning from Human Feedback (RLHF) \citep{christiano2017deep} for preference finetuning. The underlying principle \citep{ouyang2022training} is first to learn a reward function from a preference dataset and then to optimize this reward using a Reinforcement Learning (RL) approach such as policy gradient \citep{williams1991function} or Proximal Policy Optimization (PPO) \citep{schulman2017proximal}. These approaches cannot train from a fixed preference dataset alone, they also need fresh generations from the learnt model. %
\looseness=-1

More recently, direct alignment methods were introduced, such as Direct Policy Optimization (DPO) \citep{rafailov2023direct}, Identity Policy Optimization (IPO) \citep{azar2024general} or Sequence Likelihood Calibration (SLiC-HF) \citep{zhao2023slic}, among many others. 
These approaches are usually considered simpler, more stable, and computationally more efficient.
\looseness=-1

Whether classic RLHF or direct alignment, the losses or gradients involved depend on the likelihood of completion from a dataset or are generated according to the policy. The likelihood of a sequence is not length-invariant, and therefore RLHF and direct alignment are not length-invariant, departing from the cross-entropy loss used for pretraining and supervised finetuning.
\looseness=-1

We address this discrepancy in a principled way. To do so, we introduce a new averaging operator for policies and explain how to compose it with the operator providing the optimal RL solution so as to make optimization length-invariant. We also show how this general approach can be applied to direct alignment, notably demonstrating that this simply translates into replacing log-likelihoods by length-normalized log-likelihoods in the underlying loss function from first principles. 
Eventually, we study empirically the effect of such averaging, observing a trade-off between between the length of generations and their scores.

\section{Background}

We note $\x$ a prompt and $\y$ a generation or completion, and we adopt an RL viewpoint by calling the LLM to be trained a policy $\pi(\y|\x)$. The term $\y$ is a sequence of length $\ly$, $\y = (y_1, \dots, y_\ly)$, and the policy is an autoregressive model, $\pi(\y|\x) = \prod_{t=1}^\ly \pi(y_t|\x, \y_{<t})$.
For pretraining and supervised finetuning, the policy is optimized by maximizing the log-likelihood, that is, by minimizing the cross-entropy loss. For example, for supervise finetuning, the sampled-based loss for a pair $(\x,\y)$ is
\begin{equation}
    \ell_\text{xe}(\x,\y;\pi) = - \frac{1}{\ly} \sum_{t=1}^\ly \ln \pi(y_t|\x, \y_{<t}).
\end{equation}
So, it is, in practice, averaged token-wise.

\subsection{RLHF }
Assume that we have a dataset of preferences $\mathcal{D}= \{(\x,\y^+,\y^-)\}$, where completion $\y^+$ is preferred to $\y^-$ given the prompt $\x$ according to human judgement. A first step is to learn a reward model $R(\x,\y)$ from these preferences. For example, one can use a Bradley-Terry model \citep{bradley1952rank},
\begin{equation}
    \ell_\text{BT}(\x,\y^+,\y^-;R) = - \ln \sigma(R(\x,\y^+)-R(\x,\y^-)),
\end{equation}
with $\sigma$ the sigmoid function. Then, one optimizes this learned reward using an RL approach. Typically, a regularized RL \citep{fox2015taming,jaques2017sequence,geist2019theory} approach is considered, optimizing the reward while not deviating too much from some reference model $\piref$. Generally, it is the result of supervised finetuning,  also used at initialization. Formally, the corresponding objective function is, with $\rho$ a distribution over prompts:
\looseness=-1
\begin{equation}
    J(\pi) = \E_{\x\sim\rho, \y\sim\pi(\cdot|\x)}[R(\x,\y) - \beta \ln\frac{\pi(\y|\x)}{\piref(\y|\x)}].
\end{equation}
Optimizing $J$ can be achieved by using, for example, (regularized) policy gradient or PPO, both of which require fresh generations.

\subsection{Direct alignment}
Direct alignment
aims at learning the policy directly from preference data, without going through the proxy reward function. The related literature is rich \citep{rafailov2023direct,zhao2023slic,azar2024general}, we provide here a general derivation~\citep{tang2024generalized}.
\looseness=-1

Let $h$ be a convex classification function. One can use $h$ for learning a reward model by minimizing the loss (generalizing $\ell_\text{BT}$):
\begin{equation}
    \ell_h(\x,\y^+,\y^-;R) = h(R(\x,\y^+)-R(\x,\y^-)).
    \label{eq:ell_h}
\end{equation}
For example, $h(z) = -\ln\sigma(z)$ corresponds to DPO, $(\frac 1 2 -z)^2$ to IPO and $\max(0, 1-z)$ to SLiC-HF, other choices being possible. 
Then, the idea is to exploit the fact that the solution to $J(\pi)$ is known analytically:
\begin{equation}
    \pi_R(\y|\x) = \frac{\piref(\y|\x)\exp\frac{R(\x,\y)}{\beta}}{Z_R(x)},
    \label{eq:rl_optimal_solution}
\end{equation}
with $Z_R(x)$ an intractable partition function. From there, we can easily express the reward as a function of the policy:
\begin{equation}
    R(\x,\y) = \beta\ln\frac{\pi(\y|\x)}{\piref(\y|\x)} + \beta\ln Z_R(x).
    \label{eq:reward_to_policy}
\end{equation}
Then, we can inject this into $\ell_h$, and because the loss is contrastive, the intractable log-partition terms cancel out (also dropping the $R$ index):
\begin{align}
    &\ell_h(\x,\y^+,\y^-;\pi)
    \\
    = &h\left(\beta\left(\ln\frac{\pi(\y^+|\x)}{\piref(\y^+|\x)} - \ln\frac{\pi(\y^-|\x)}{\piref(\y^-|\x)}\right)\right).
\end{align}
Thus, optimizing this loss allows finetuning the LLM directly from preference data without learning a proxy reward or using an RL algorithm.

However, be it RLHF or direct alignment, related losses all involve log-likelihood rather than averaged log-likelihood as in the cross-entropy loss $\ell_\text{xe}$. These losses are not linear in the log-likelihood (contrary to the cross-entropy), and as they involve completions of possibly different lengths, how to soundly introduce averaging is not clear.

\section{Averaging log-likelihoods}

\subsection{General approach }
Let $\pi$ be any policy. We introduce an averaging operator $F:\pi \rightarrow \pi_F$, that transforms a policy $\pi$ into a policy $\pi_F$ defined as
\begin{align}
    \pi_F(\y|\x) &= \frac{(\pi(\y|\x))^{1/\ly}}{Z_{\pi,F}(\x)} 
    = \frac{\sqrt[\ly]{\prod_{t=1}^\ly \pi(y_i|\x,\y_{<i}})}{Z_{\pi,F}(\x)},
\end{align}
with $Z_{\pi,F}$ a partition function. So, we define the sequence probability of $\pi_F$ to be the geometric mean of the token probabilities of $\pi$. This amounts to averaging the log-likelihood, up to a shift by the related partition function:
\begin{equation}
    \ln\pi_F(\y|\x) = \frac{1}{\ly} \ln \pi(\y|\x) - \ln Z_{\pi,F}(x).
\end{equation}
Notice that $F$ is a bijection, and $F^{-1}$ applied to $\pi$ gives a policy $\pi_{F^{-1}}(\y|\x)\propto \pi(\y|\x)^\ly$.

How to use this in an RL context? Let $T_R:\pi\rightarrow \pi_{T_R}$ be the operator that transforms a policy $\pi$ into $\pi_{T_R}$, defined as the solution to the RL problem when regularized towards $\pi$, that is 
$\pi_{T_R}(\y|\x) \propto \pi(\y|\x) \exp\frac{R(\x,\y)}{\beta}$.
With this notation, the classic solution of the RL problem regularized towards $\piref$ depicted in Eq.~\eqref{eq:rl_optimal_solution} can be written as $\pi_R = T_R \piref$.

For accounting for length normalization in an RL context, we propose the following solution:
\looseness=-1
\begin{equation}
    \tilde{\pi}_R = F^{-1} T_R F \piref.
\end{equation}
In other words, we first length-normalize the reference policy using the operator $F$, giving $F \piref$. Then, we find the optimal policy using the operator $T_R$ for the length-normalized reference $F\piref$. Lastly, we reverse the length normalization using the inverse operator $F^{-1}$ to go back to the original space. Therefore, in some sense, the optimization process (through operator $T_R$) is length-agnostic because it is applied to $F\piref$, while the final solution is length-aware thanks to using the operator $F^{-1}$.

\subsection{Application to direct alignment }
There remains to know how to get a practical algorithm from these derivations, especially given that the associated partition functions are intractable. We illustrate that in the case of direct alignment methods. The key to deriving them is Eq.~\eqref{eq:reward_to_policy}, which is nothing else than a rewriting of $\pi = T_R \piref$. We propose to simply do the same thing with $\pi = F^{-1} T_R F \piref \Leftrightarrow F\pi = T_R F \piref$. This is equivalent to
\begin{equation}
    \frac{(\pi(\y|\x))^{1/\ly}}{Z_1(x)} = \frac{\frac{(\piref(\y|\x)^{1/\ly})}{Z_2(x)}\exp\frac{R(\x,\y)}{\beta}}{Z_3(x)}, 
\end{equation}
where the $Z_i$'s are the partition functions associated with each operator. Writing $Z = \frac{Z_3 Z_2}{Z_1 }$, taking the logarithm and averaging, we obtain:
\begin{equation}
    R(\x,\y) = \beta \frac{1}{\ly}\ln \frac{\pi(\y|\x)}{\piref(\y|\x)} + \beta \ln Z(x).
\end{equation}
Hence, injecting this into $\ell_h$ in Eq.~\eqref{eq:ell_h} gives (thanks to the terms $\ln Z(x)$ cancelling again):
\begin{GrayBox}
\vspace{-1em}
\small
\begin{align}
    &\ell^\text{avg}_h(\x,\y^+,\y^-;\pi) =
    \\
    &h\left(\beta\left(\frac{1}{\lyp}\ln\frac{\pi(\y^+|\x)}{\piref(\y^+|\x)} - \frac{1}{\lym}\ln\frac{\pi(\y^-|\x)}{\piref(\y^-|\x)}\right)\right).
\end{align}
\end{GrayBox}
So, in the case of direct alignment, the proposed length-agnostic RL approach very simply amounts to normalizing each involved log-likelihood by the corresponding length.
\looseness=-1

\section{Related works}

There is a large literature on direct alignment, \textit{eg.} \citet{rafailov2023direct,zhao2023slic,azar2024general}. To the best of our knowledge, most of these approaches do not average likelihoods. A notable exception is RRHF (Rank Responses to align Human Feedback) \citep{yuan2023rrhf}, which uses normalized log-likelihood in a ranking loss. However, it does so without justification and does not compare to not normalizing.

The implementation may sometimes differs from what is presented in a research article. For example, IPO \citep{azar2024general} was introduced with only experiments on bandits, not on LLMs, and without discussing the fact of averaging or not log-likelihoods. However, the IPO implementation in HugginFace\footnote{\url{https://github.com/huggingface/trl/blob/main/trl/trainer/dpo_trainer.py}} does perform such an averaging\footnote{\url{https://github.com/huggingface/trl/pull/1265}}.

To our knowledge, we are the first to formalize mathematically the averaging log-likelihoods systematically for direct alignment and to provide an empirical comparison of averaging them or not.

\section{Experiments}

\begin{figure*}[tbh]
    \centering
    \begin{minipage}{.315\linewidth}
        \centering
        \includegraphics[width=\linewidth]{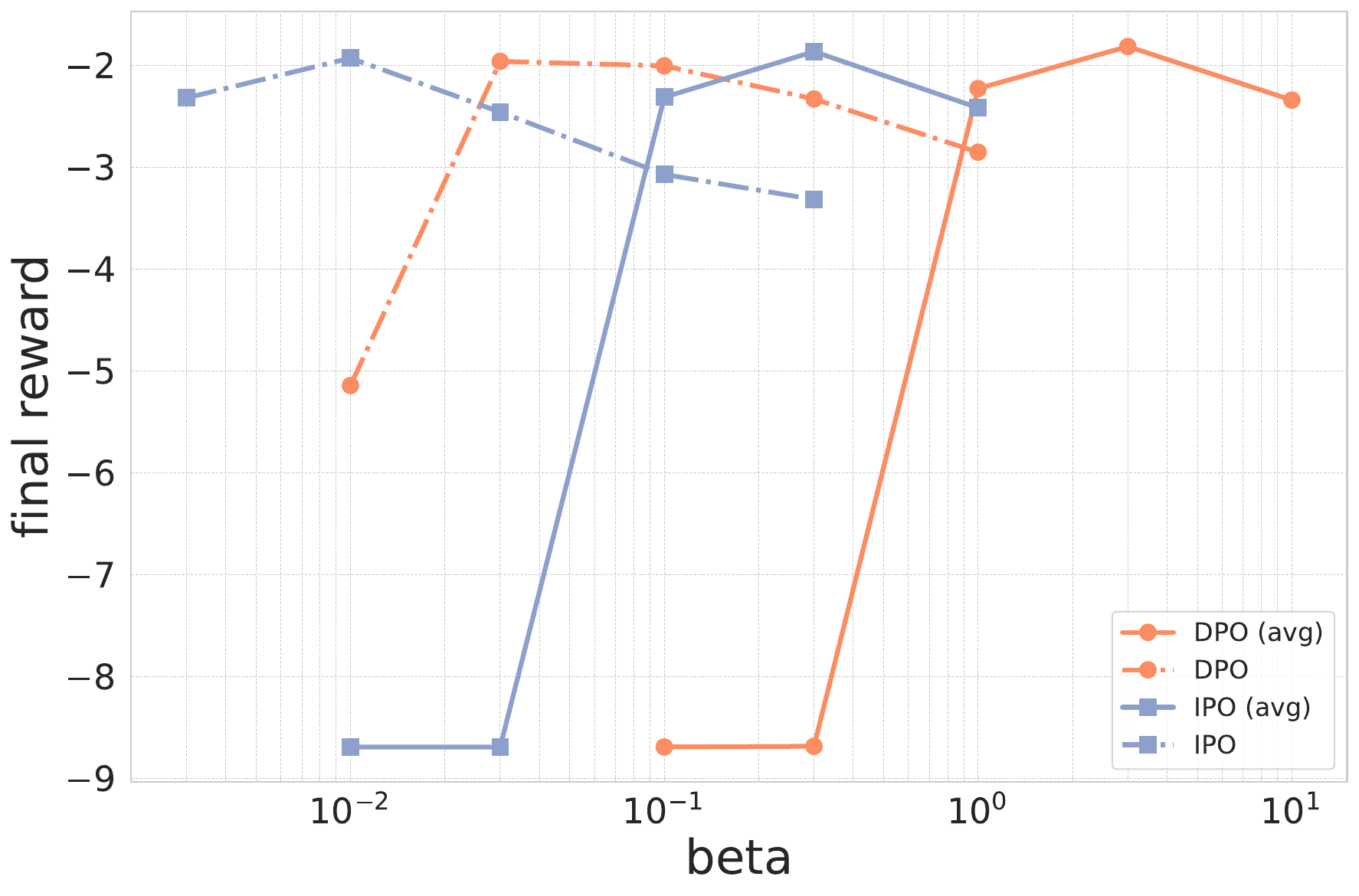}
        \caption{Final reward as a function of the temperature $\beta$.}
        \label{fig:avg:reward_beta}
    \end{minipage}%
    \hfill
    \begin{minipage}{.315\linewidth}
        \centering
        \includegraphics[width=\linewidth]{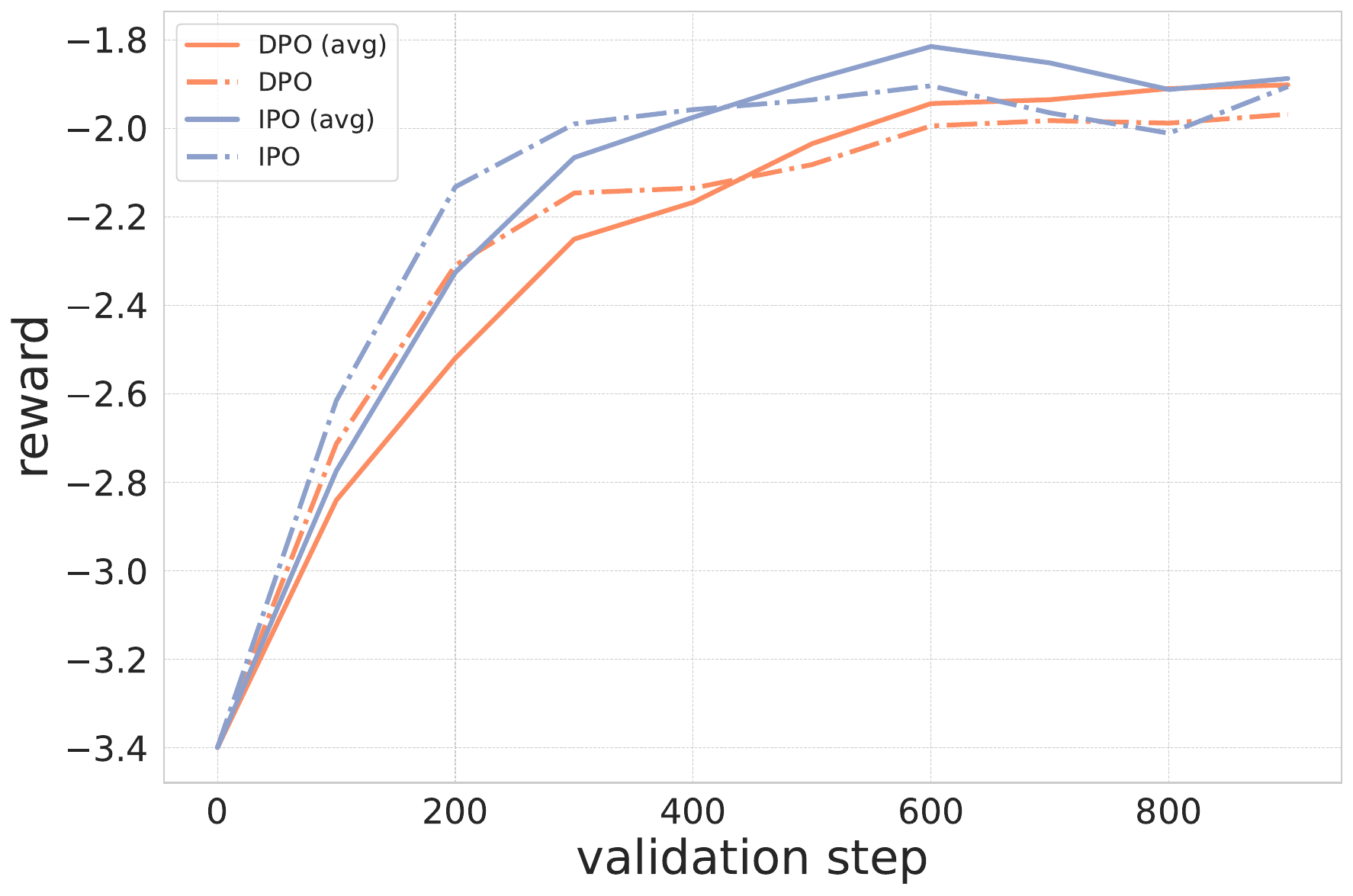}
        \caption{Average reward of generations during training.}
        \label{fig:avg:reward}
    \end{minipage}%
    \hfill
    \begin{minipage}{.315\linewidth}
        \centering
        \includegraphics[width=\linewidth]{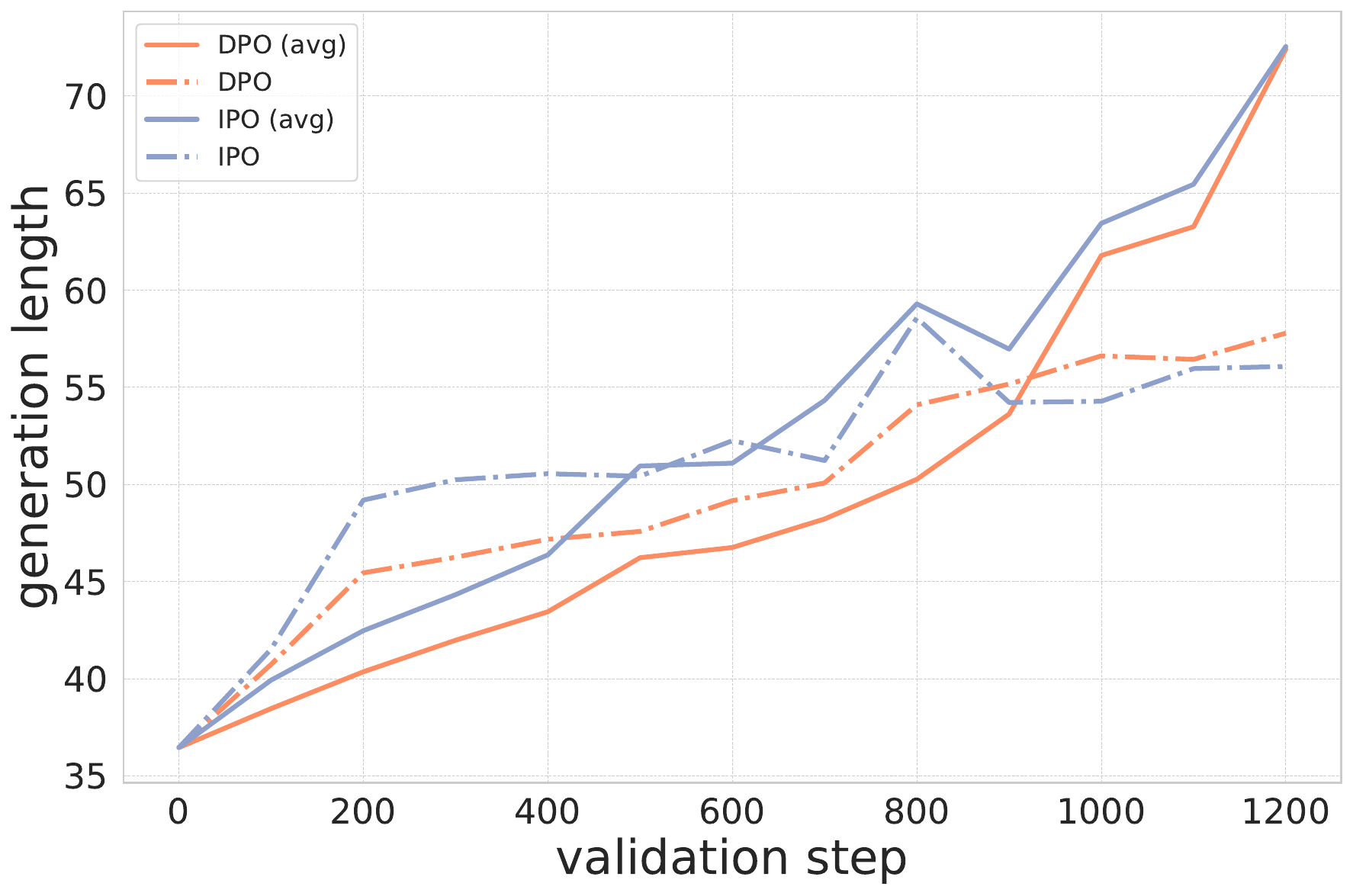}
        \caption{Average length of generations during training.}
        \label{fig:avg:length}
    \end{minipage}
\end{figure*}

Our objective is to study the effect of averaging the likelihoods in direct alignment, and we will consider, more specifically, DPO, IPO, and their averaging counterparts in optimizing the preferences of a summarization task. We provide a short description of the setting. All missing experimental details are provided in Appx.~\ref{appx:exp_details}.   

We consider a summarization task by using the Reddit TL;DR dataset \citep{stiennon2020learning}. As a policy model, we use Llama2-7B \citep{touvron2023llama2}, finetuned on the SFT train split. We use it to train a reward model (RM) on the train split of the dataset. Even if we do direct alignment, which does not require a reward model, we use it as ground truth for both ranking the completions in the dataset (accuracy of $89.1\%$ on the train set and $72.1\%$ on the validation set), and for evaluating generations. That is, we use the reward model as a convenient proxy for evaluation. We do not claim it to be the best signal to optimize, but this is a relevant metric for this study.

We train IPO, DPO, and their averaging counterparts, for 2 epochs. We perform 128 generations every 100 training steps and score them according to the RM. Fig.~\ref{fig:avg:reward_beta} shows the final reward for each approach as a function of $\beta$. We observe that DPO obtains its best result for higher values of $\beta$, and that averaging counterparts obtain their best value for ten times the original $\beta$.

Then, we train again each method for its best value of $\beta$, and perform 1024 generations every 100 training steps. We compute the average reward and length of each of these generations, the results are presented in Figs.~\ref{fig:avg:reward} and~\ref{fig:avg:length}. On the reward side, we observe a similar behavior for both DPO and IPO, the reward of the averaging counterparts increases more slowly, then reaches higher values, before stabilizing to a similar reward. On the length side, generations of the averaging counterparts are first shorter, before reaching a higher peak at two epochs. This is suspicious, a possible sign of reward hacking, and we checked the generation manually. We indeed observe a phenomenon of reward hacking for all approaches, but it is more marked for averaging counterparts, with sequences ending with never-ending punctuation. We provide such examples in Appx.~\ref{appx:gen_examples}.

\begin{figure}
    \centering
    \includegraphics[width=\linewidth]{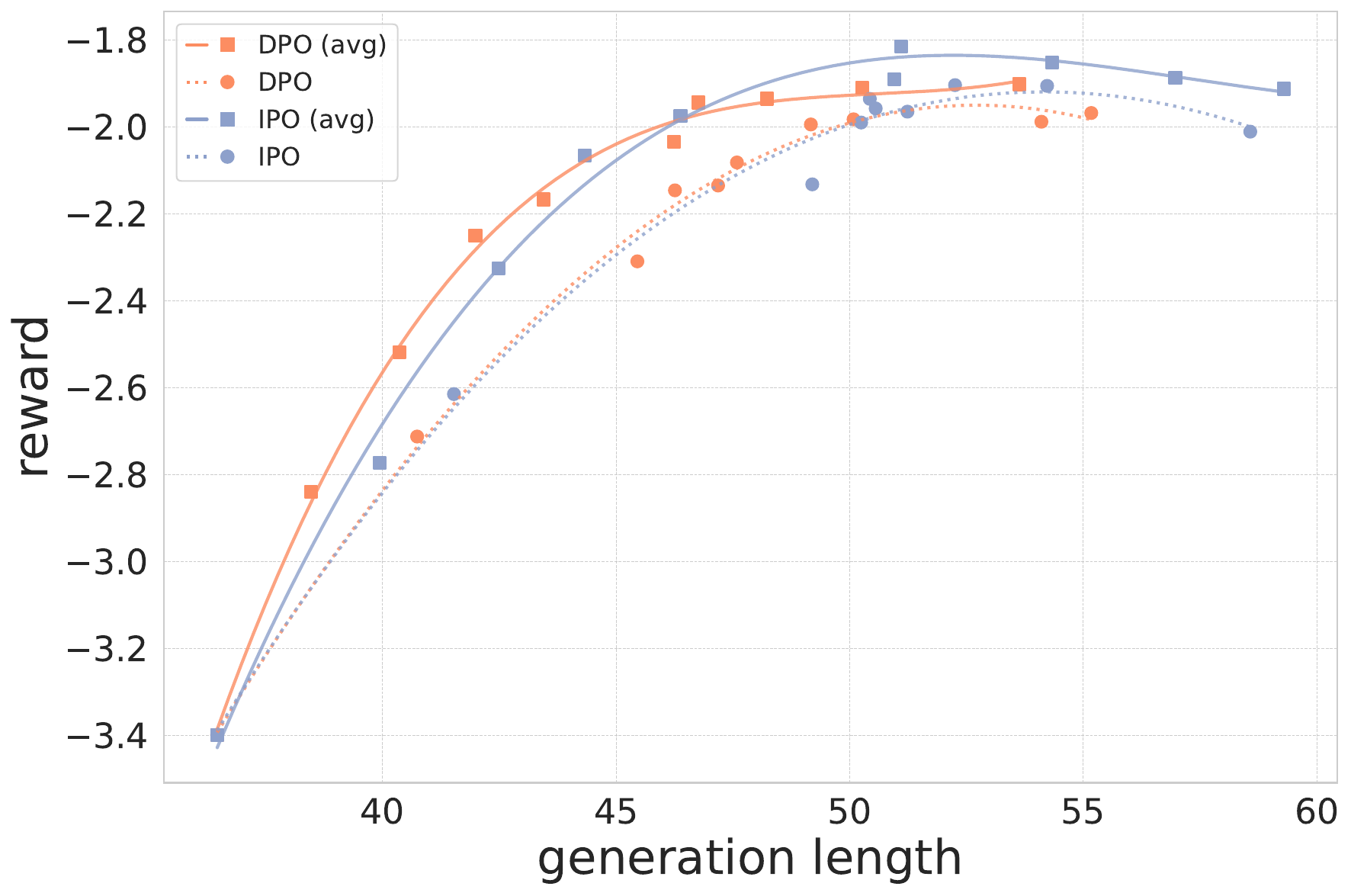}
    \caption{Relationship between lengths and rewards.}
    \label{fig:avg:length_reward}
\end{figure}

To get a better sense of the link between the length and reward of generations, Fig.~\ref{fig:avg:length_reward} provides a scatter plot of rewards as a function of lengths (each point corresponding to the 1024 generations of one of the validation steps for one the methods), with a polynomial interpolation. To remove the reward hacking regime discussed above, we only consider the data of the 1.5 first epochs.

We can observe that the Pareto front of the averaging counterparts (plain line) is above that of the standard approaches (dashed line), meaning that the averaging counterparts of DPO and IPO achieve higher rewards for shorter completions. This exhibits an interesting gain in the trade-off between length and quality (even though measured through the reward model) of generations. We also hypothesize that averaging the likelihood could make the choice of $\beta$ more invariant to the dataset, especially with varying lengths of completions, but this experiment does not provide empirical evidence on this.
\looseness=-1

\section{Discussion and perspectives}

We have proposed a systematic and mathematically principled approach for replacing likelihoods with averaged likelihoods in direct alignment methods. This kind of approach could be more broadly applicable to RLHF, which is an interesting future research direction. 
We have compared averaging the likelihood or not with two classic direct alignment approaches, DPO and IPO. If, in our experiments, averaging the likelihood seems to be more prone to reward hacking or overfitting, it also provides a sensible gain in the trade-off between the length and quality of the generations, as illustrated Fig.~\ref{fig:avg:length_reward}. We plan to consolidate these results on more tasks and with direct alignment methods. 
Overall, length normalization in direct alignment is very simple, and we hope that our mathematical justification will help make it a standard tool.
\looseness=-1

\section*{Limitations}
Here, we highlight some of this method's limitations. The proposed approach to making direct alignment independent of generation length is evaluated using a reward model, which, while convenient, may not fully capture the breadth of human preferences. In addition, the reward hacking observed during the experiment, which we dealt with using early stopping, highlights areas of potential refinement such as regularization to avoid undesirable generation patterns. Finally, the computational cost associated with training and evaluating such LLMs limits the scope of empirical validation.

\bibliography{custom}

\newpage

\appendix

\onecolumn

\section{Experimental details}
\label{appx:exp_details}

This appendix provides details on the experimental and training.

\textbf{Dataset.} We use the Reddit TL;DR dataset\footnote{\url{https://github.com/openai/summarize-from-feedback}} from \citet{stiennon2020learning}. This summarization dataset has an SFT split (prompts being Reddit posts and completions consisting of human-written summaries) and a preference split (similar prompts and completions made of human-ranked preference pairs).

\textbf{Policy Model.} We use Llama2-7B as the base model\footnote{\url{https://huggingface.co/meta-llama/Llama-2-7b-hf}} \citep{touvron2023llama2}. We supervise finetune it on the SFT split of the TL;DR dataset, using it as an initial model for both the reward and the direct alignment approaches. This model is trained for 2 epochs with Adam, with a cosine decay scheduler ($2.10^{-5}$ to 0), warmup of $10\%$, using a batch of size 128.

\textbf{Reward Model.} As explained in the main text, even though we consider direct alignment, which does not require a reward model, we train one for both ranking the completions and for evaluation of the generations.  Again, we insist that we do not claim such a reward model to be the best thing to optimize for improving the LLM; we use it as a convenient proxy for assessing what can bring the fact of averaging the log-likelihoods. For training the reward function, we use a classic Bradley-Terry model \citep{bradley1952rank}, that is Eq.~\eqref{eq:ell_h} with $h(z) = -\ln\sigma(z)$.
The reward model is trained for two epochs on the train split of the preference dataset, with Adam, a learning rate of $10^{-6}$, a batch of size 128, and a warm-up of $10\%$ of the total number of training steps. The trained reward model achieves an accuracy of $89.1\%$ on the train set and of $72.8\%$ on the validation set.

\textbf{Training details. } We train all algorithms for two epochs over the train split of the preference dataset, using the same hyper-parameters as for the reward model (Adam with a learning rate of $10^{-6}$, batch size 128, warm-up of $10\%$). We sweep over the following values of $\beta$ for producing Fig.~\ref{fig:avg:reward_beta}:
\begin{itemize}
    \item DPO: $\beta\in\{0.01, 0.03, 0.1, 0.3, 1\}$
    \item DPO(avg): $\beta\in\{0.1, 0.3, 1, 3, 10\}$
    \item IPO: $\beta\in\{0.003, 0.01, 0.03, 0.1, 0.3\}$
    \item IPO(avg): $\beta\in\{0.01, 0.03, 0.1, 0.3, 1\}$ 
\end{itemize}
The best values of $\beta$ according to Fig.~\ref{fig:avg:reward_beta} are $\beta=0.1$ for DPO, $\beta = 3$ for DPO(avg), $\beta=0.01$ for IPO and $\beta = 0.3$ for IPO(avg).

\clearpage
\section{Generation examples}
\label{appx:gen_examples}

We list two examples from the TL;DR dataset during evaluation. We observe that the summary indeed improves in quality during training, with more elaborate structures. Besides, the summary length increases as the LLM generates additional sentences in late training. Finally, we started noticing some reward hacking artifacts after 2epoches (2x600steps) of the training (in red).

\begin{tcolorbox}
\underline{SUBREDDIT}: relationships\\
\underline{TITLE}: (24/F) I hate losing arguments to my bf (27/m)\\
\underline{POST}: I just got into another heated petty argument with my bf of 3 years! He started telling our roommate a story about someone, I was sure he was thinking of something else, he stops dead in his tracks and tells me that that's not what happened. I said more details of what I remembered, he told me again that I was wrong and I should believe him because it's a story that happened to him. He said what he always does, "why is it that when we disagree you assume I must be wrong? Why don't you ever consider that you could be wrong?" 

By that point I had realized it was possible I was wrong, but I absolutely did not want to concede that to him or offer him any slack. I don't get like this with everybody, but my bf and I both have this stupid trait of never backing down! And I hate admitting I'm wrong to him because I feel like he acts super smug and condescending, when I make a mistake he laughs at me in an "oh, you" sort of way that INFURIATES me. I know I overreact to this sort of thing but I don't know how to improve!
\end{tcolorbox}

\textbf{DPO:}
\begin{itemize}
    \item \emph{Step=300:} I hate admitting to my bf that I'm wrong and I get super defensive when he acts like he's so happy and smug that I was wrong. How do I improve?
    \item \emph{Step=600:} I hate admitting I'm wrong to my bf because he acts super smug and I don't like it. How do I get better about this?
    \item \emph{Step=900:} I hate admitting I'm wrong to my bf in arguments because he acts super smug and I can't stand it, but I also hate losing arguments so I always fight to the death with him even when I'm sure I'm wrong sometimes! What do I do?!
    \item \emph{Step=1200:} I hate admitting I'm wrong to my bf because he acts super smug and I want to win every argument. How do I improve \textcolor{red}{?!?!?!?!?!?!?!?!?!?!?!?!?!?!?!?!?!?!?!?!?!?!?!?!?!?!?!?!?!?!?!?!?!?!?!?!?!?!?!?!?!?!?!?!?!?!?!?!!?!?!?!\\?!?!?!?!?!?!?!?!?!?!?!?!?!?!?!?!?!?!?!?!?!?!?!?!!?!?!?!?!?!?!?!?!?!?!?!?!?!?!?!?!?!?!?!?!?!?!!?!?!?!?!\\!?!?!?!?!?!?!?!?!?!?!?!?!?!?!?!?!?!?!?!?!?!?!?!?!?!?!?!?!?!?!?!?!?!?!?!?!?!?!?!?!?!?!?!?!?!?!?!}
\end{itemize}   

\textbf{DPO (Avg):}
\begin{itemize}
    \item \emph{Step=300:} Girlfriend of two years dumps me unexpectedly, moves on less than two days later, leaving me completely lost. Where do I start to move on from something like that?
    \item \emph{Step=600:} Girlfriend of two years dumps me unexpectedly, moves on very quickly, and it really hurts. I need help moving on because I'm lost.
    \item \emph{Step=900:} My boyfriend wants to experience other girls after me (his first girlfriend), I said i'm down for ffm but I feel really insecure about that now and I don't know what to do. I really want to stay with him.
    \item \emph{Step=1200:} I hate backing down in arguments with my bf and I hate even more when he acts smug and condescending after he wins an argument! I want to improve how I react but I don't know how?!
\end{itemize}   

\textbf{IPO:}
\begin{itemize}
    \item \emph{Step=300:} I hate admitting to my bf that I'm wrong and I get super defensive when he acts like he's so happy that he was right and I was wrong. I want to improve so I don't get so upset over this!
    \item \emph{Step=600:} I hate admitting I'm wrong to my bf because he acts super smug and I don't like it, but I also hate losing arguments and always want to be right. I need to learn how to not care so much!
    \item \emph{Step=900:} I hate admitting I'm wrong to my bf in arguments because he acts super smug and I can't stand it, but I also hate losing arguments! How do I improve so I don't feel so defensive about being wrong?
    \item \emph{Step=1200:} I hate admitting I'm wrong to my bf in arguments because he acts super smug and I want to win every argument anyway even if I'm not 100\% sure I'm right. How do I stop?
\end{itemize}   

\textbf{IPO (Avg):}
\begin{itemize}
    \item \emph{Step=300:} Girlfriend of two years dumps me unexpectedly, moves on less than two days later, leaving me completely lost. Where do I start to move on from something I've revolved my entire life around?
    \item \emph{Step=600:} I hate admitting I'm wrong to my bf because he acts super smug and I am terrible at backing down in arguments. How do I improve?!?!
    \item \emph{Step=900:} I hate admitting defeat in petty arguments to my bf, he always acts so smug and I want to know how to improve so I can save some face!
    \item \emph{Step=1200:} My bf and I refuse to back down in arguments and I hate losing them. He has this annoying way of making me feel stupid when I admit I was wrong and I hate it. How do I improve\textcolor{red}{?!?!?!???????????}
\end{itemize}

\vspace{3em}
\begin{tcolorbox}
\underline{SUBREDDIT:} relationships \\
\underline{TITLE:} Me [19 F] with my boyfriend [19 M ] 3.5 years, when is it time to let go? with line breaks. \\
\underline{POST:} Okay, so I've never really posted anything before, so sorry if this is weird or I'm not doing it right or something. I started dating my current BF when I was 15 and a sophomore in high school, he was 16 and a junior. Now we are both in college. We are insanely close. Like spend literally ALL our time together. He is incredibly important to me and we do everything together. I love him a lot and I have incredible respect for him as a person. 
The problem is we both don't have friends outside of the relationship. 
I mean we both have friends. We have a lot of friends really, but we just don't feel like we connect with anyone other than each other anymore. It's hard to explain. 
We both ditched all of our friends from high school. Some of it was because of our relationship, but a lot of it was high school is shitty. 
\end{tcolorbox}

\newpage

\textbf{DPO:}
\begin{itemize}
    \item \emph{Step=300:} I know I need to break up with my boyfriend of 3.5 years, but I'm scared because I feel like I won't ever connect with anyone again because we have kind of abandoned our friends for each other.
    \item \emph{Step=600:} I know I need to break up with my boyfriend of 3.5 years, but I feel like I won't ever connect with anyone else because I've been so focused on this relationship my whole life. I don't know what to do.
    \item \emph{Step=900:} I know I need to break up with my boyfriend I've been with for 3.5 years, but I'm afraid because I feel like I won't ever connect with anyone again after him. I feel like I only connect with him. I don't feel like I connect with anyone else.
    \item \emph{Step=1200:} My boyfriend and I have pretty much zero outside friends and we are too dependent on each other, I know we should break up but I'm scared I will never connect with anyone else. I'd really appreciate some advice. Thanks.
\end{itemize}   

\textbf{DPO (Avg):}
\begin{itemize}
    \item \emph{Step=300:} I know I need to break up with my boyfriend of 3.5 years but I feel like I won't ever connect with anyone again if I do.
    \item \emph{Step=600:} I know I need to break up with my boyfriend but I feel like I won't have any friends after because I only really connect with him. Anybody else experience something like this?</s>
    \item \emph{Step=900:} I know its time to move on from my long term relationship, but I'm scared because I feel like I won't ever really connect with anyone anymore if I do. (Friends outside of the relationship.)
    \item \emph{Step=1200:} My boyfriend and I have pretty much abandoned our friendships for each other and I know that I need to break up with him, but I feel like I won't ever connect with anyone outside of this relationship anymore and I'm terrified of losing friends. Has anyone else experienced this? What do I do?\textcolor{red}{***---}
\end{itemize}   

\textbf{IPO:}
\begin{itemize}
    \item \emph{Step=300:} I know I need to break up with my boyfriend of 3.5 years, but I'm scared because I feel like I won't ever connect with anyone outside of my SO now. Has anyone expressed something like this?
    \item \emph{Step=600:} I know I need to end my relationship with my boyfriend whom I love very dearly, but I'm scared because I feel like I won't ever connect with anyone outside of the relationship anymore if I do.
    \item \emph{Step=900:} I know its time to move on from my long term relationship, but I'm scared because I don't feel like I'll ever connect with anyone again outside of my SO. **Has anyone else expressed this feeling before?** If so what did you do?
    \item \emph{Step=1200:} My boyfriend and I have pretty much ditched all of our friends from high school and replaced them with each other. We know we need to break up eventually, but I'm scared because I don't feel like I'll ever connect with anyone else. Has anyone else experienced something like this?
\end{itemize}   

\newpage
\textbf{IPO (Avg):}
\begin{itemize}
    \item \emph{Step=300:} Girlfriend of two years dumps me unexpectedly, moves on less than two days later, leaving me completely lost. Where do I start to move on from something I've revolved my entire life around?
    \item \emph{Step=600:} I know I need to end my relationship with my boyfriend whom I love very dearly, but I am afraid because I feel like I won't ever connect with anyone else outside of this relationship because my boyfriend is my main group of friends now.
    \item \emph{Step=900:}I know its time to move on from my long term relationship, but I'm scared because I feel like I won't ever find anyone I connect with again. Anybody relate? Or have anything to contribute? Thanks :)
    \item \emph{Step=1200:} My boyfriend and I have pretty much abandoned our friendships for each other and I know the relationship is overdue for us both having bekend break from all this tunnel vision but I'm scared I won't find any new friends that I actually enjoy being around. Has anyone else experienced this? What do I do?\textcolor{red}{***}
\end{itemize}

\end{document}